\documentclass{sig-alternate}

\usepackage{graphicx}
\usepackage{amsmath,amssymb} 
\usepackage{color}

\usepackage{cite}
\graphicspath{{./images/}}

\newcommand{\hide}[1]{}

\newcommand{\etal}{\textit{et al.}~}

\begin{document}






%

\title{Compact Hash Codes for Efficient Visual Descriptors Retrieval in Large Scale Databases}
%
%
%
%
%

\numberofauthors{1} 
%
\author{
%
%
\alignauthor{Simone Ercoli, Marco Bertini and Alberto Del Bimbo}\\
\affaddr{Media Integration and Communication Center, Universit\`a degli Studi di Firenze}\\
\affaddr{Viale Morgagni 65 - 50134 Firenze, Italy}\\
\email{ \{\textit{name.surname}\}@unifi.it }
}

\date{30 July 1999}

\maketitle
\begin{abstract}
In this paper we present an efficient method for visual descriptors retrieval based on compact hash codes computed using a multiple k-means assignment. The method has been applied to the problem of approximate nearest neighbor (ANN) search of local and global visual content descriptors, and it has been tested on different datasets: three large scale public datasets of up to one billion descriptors (BIGANN) and, supported by recent progress in convolutional neural networks (CNNs), also on the CIFAR-10 and MNIST datasets. Experimental results show that, despite its simplicity, the proposed method obtains a very high performance that makes it superior to more complex state-of-the-art methods.
\end{abstract}

%
%


%
%

%
%



\section{Introduction}\label{sec:intro}

Efficient nearest neighbor (NN) search is one of the main issues in large scale information retrieval for multimedia and computer vision tasks. When dealing with high dimensional features also methods for multidimensional indexing obtain performance comparable to that of exhaustive search \cite{jegou-2011}. A typical solution is to employ methods that perform approximate nearest neighbor (ANN) search, typically using feature hashing and Hamming distances. 
These methods generally use inverted files, e.g.~implemented using hash tables, and require to use hash codes with a length of several tens of bits to obtain a reasonable performance in retrieval. This combination requires quite large amounts of memory to store large scale databases of features, thus their application to systems with relatively limited memory (e.g.~mobile devices still have 1-2 GB RAM only) or systems that involve a large-scale media analysis is not feasible.

In this paper we present a novel method for feature hashing, based on multiple k-means assignments, that is unsupervised, that requires a very limited codebook size and that obtains good performance in retrieval even with very compact hash codes. The proposed approach greatly reduces the need of training data and memory requirements for the quantizer, and obtains a retrieval performance similar or superior to more complex state-of-the-art approaches on standard large scale datasets. This makes it  suitable, in terms of computational cost, for mobile devices, large-scale media analysis and retrieval in general.

\section{Previous Works}\label{sec:previous-work}

Previous works on visual feature hashing can be divided in methods based on hashing functions, scalar quantization, vector quantization and, more recently, neural networks.

\paragraph{Hashing functions} 
Weiss \etal\cite{weiss-2009} have proposed to treat the problem of hashing as a particular form of graph partitioning, in their Spectral Hashing algorithm.

Heo \etal\cite{heo-2012} have proposed to encode high-dimensional data points using hyperspheres instead of hyperplanes; Jin \etal\cite{jin-2014} have proposed a variation of LSH, called Density Sensitive Hashing, that does not use random projections but instead uses projective functions that are more suitable for the distribution of the data. 

Du \etal\cite{du-2014} have proposed the use of Random Forests to perform linear projections, along with a metric that is not based on Hamming distance.

Paulev\'e \etal\cite{pauleve-2010} have compared structured quantization algorithms with structured quantizers (i.e.~k-means and hierarchical k-means clustering). Experimental results on SIFT descriptors have shown that unstructured quantizers provide significantly superior performances  with respect to structured quantizers.

\paragraph{Scalar Quantization} 
Zhou \etal\cite{zhou-2012} have proposed an approach based on scalar quantization of SIFT descriptors. The median and the third quartile of the bins of the descriptor are computed and used as thresholds, hashing is then computed coding the value of each bin of the descriptor with 2 bits, depending on this subdivision. The final hash code has a dimension of 256 bits, but only the first 32 bits are used to index the code in an inverted file.
The method of \cite{zhou-2012} has been extended by Ren \etal\cite{ren-2014}, including an evaluation of the reliability of bits, depending on their quantization errors. Unreliable bits are then flipped when performing search, as a form of query expansion. 
Chen and Hsieh \cite{chen-2015} have recently proposed an approach that quantizes the differences of the bins of the SIFT descriptor, using the median computed on all the SIFT descriptors of a training set as a threshold.

\paragraph{Vector Quantization} 
J\'egou \etal\cite{jegou-2011} have proposed to decompose the feature space into a Cartesian product of subspaces with lower dimensionality, that are quantized separately. This Product Quantization (PQ) method is efficient in solving memory issues that arise when using vector quantization methods such as k-means, since it requires a much reduced number of centroids. The method has obtained state-of-the-art results on a large scale SIFT features dataset, improving over methods such as SH \cite{weiss-2009} and Hamming Embedding \cite{jain-2011}. This result is confirmed in the work of Chandrasekhar \etal\cite{chandrasekhar-2010}, that have compared several compression schemes for SIFT features. 

The success of the Product Quantization method has led to development of several variations and improvements.
The idea of compositionality of the PQ approach has been further analyzed by Norouzi and Fleet \cite{norouzi-2013}, that have built upon it proposing two variations of k-means: Orthogonal k-means and Cartesian k-means.
Also Ge \etal\cite{ge-2013} have proposed another improvement of PQ, called OPQ, that minimizes quantization distortions w.r.t.~space decomposition and quantization codebooks; He \etal\cite{he2013k} have proposed an affinity-preserving technique to approximate the Euclidean distance between codewords in k-means method.
Kalantidis and Avrithis  \cite{kalantidis2014locally} have presented a simple vector quantizer (LOPQ) which uses a local optimization over a rotation and a space decomposition and apply a parametric solution that assumes a normal distribution.

Babenko and Lempitsky \cite{babenko-2012} have proposed an efficient similarity search method, called inverted multi-index (Multi-D-ADC); this approach generalizes the inverted index by replacing vector quantization inside inverted indices with product quantization, and building the multi-index as a multi-dimensional table. 

\paragraph{Neural Networks}  
Lin \etal\cite{lin2015deep} have proposed a deep learning framework to create hash-like binary codes for fast image retrieval. Hash codes are learned in a point-wise manner by employing a hidden layer for representing the latent concepts that dominate the class labels (when the data labels are available). This layer learns specific image representations and a set of hash-like functions.

Do \etal\cite{do2015discrete} have addressed the problem of learning binary hash codes for large scale image search using a deep model which tries to preserve similarity, balance and independence of images. Two sub-optimizations during the learning process allow to efficiently solve binary constraints.

Guo and Li \cite{guo2015cnn} have proposed a method to obtain the binary hash code of a given image using binarization of the CNN outputs of a certain fully connected layer.

Zhang \etal\cite{zhang2015supervised} have proposed a very deep neural networks (DNNs) model for supervised learning of hash codes (VDSH). They use a training algorithm inspired by alternating direction method of multipliers (ADMM) \cite{boyd2011distributed}. The method decomposes the training process into independent layer-wise local updates through auxiliary variables.

Xia \etal\cite{xia2014supervised} have proposed an hashing method for image retrieval which simultaneously learns a representation of images and a set of hash functions.

\section{The Proposed Method}\label{sec:method}

The proposed method exploits a novel version of the k-means vector quantization approach, introducing the possibility of assignment of a visual feature to multiple cluster centers during the quantization process. This approach greatly reduces the number of required cluster centers, as well as the required training data, performing a sort of quantized codebook soft assignment for an extremely compact hash code of visual features.

The first step of the computation is a typical k-means algorithm for clustering. Given a set of observations $(\textbf{x}_1,\textbf{x}_2, \ldots, \textbf{x}_n,)$ where each observation is a d-dimensional real vector, k-means clustering partitions the $n$ observations into $k (\leq n)$ sets $\textbf{S} = \left\{S_1, S_2, \ldots, S_k\right\} $ so as to minimize the sum of distance functions of each point in the cluster to the $C_k$ centers. Its objective is to find:

\begin{equation}\label{eq:kmeans1}
argmin_\textbf{s}\sum_{i=1}^k\sum_{\textbf{x}\in S_i}\parallel   \textbf{x}-C_i\parallel^2
\end{equation}

%
%
%
%
%
%
%
%
%


This process is convergent (to some local optimum) but the quality of the local optimum strongly depends on the initial assignment. We use the \textit{k-means++} \cite{arthur2007k} algorithm for choosing the initial values, to avoid the poor clusterings sometimes found by the standard k-means algorithm.

\subsection{Multi-k-means Hashing}\label{sec:hashing}
K-means is typically used to compute the hash code of visual feature in unstructured vector quantization, because it minimizes the quantization error by satisfying the two Lloyd optimality conditions \cite{jegou-2011}. In a first step a dictionary is learned over a training set and then hash codes of features are obtained by computing their distance from each cluster center. Vectors are assigned to the nearest cluster center, whose code is used as hash code. Considering the case of 128-dimensional visual content descriptors like SIFT or the FC7 layer of the VGG-M-128 CNN \cite{Chatfield14}, this means that compressing them to 64 bits codes requires to use $k=2^{64}$ centroids. In this case the computational cost of learning a k-means based quantizer becomes expensive in terms of memory and time because: \emph{i)} there is the need of a quantity of training data that is several times larger than $k$, and \emph{ii)} the execution time of the algorithm becomes unfeasible.
Using hierarchical k-means (HKM) makes it possible to reduce execution time, but the problem of memory usage and size of the required learning set affect also this approach.
Since the quantizer is defined by the $k$ centroids, the use of quantizers with a large number of centroids may not be practical or efficient: if a feature has a dimension $D$, there is need to store $k \times D$ values to represent the codebook of the quantizer.
A possible solution to this problem is to reduce the length of the hash signature, but this typically affects negatively retrieval performance.
The use of product k-means quantization, proposed originally by J\'egou \etal\cite{jegou-2011}, overcomes this issue. 

In our approach, instead, we propose to compute a sort of soft assignment within the k-means framework, to obtain very compact signatures and dimension of the quantizer, thus reducing its memory requirements, while maintaining a retrieval performance similar to that of \cite{jegou-2011}.

The proposed method, called \textit{multi-k-means}, starts learning a standard k-means dictionary as shown in Eq.~\ref{eq:kmeans1}, using a very small number $k$ of centroids to maintain a low computational cost.
Once we obtained our $\textbf{C}_1, \ldots, \textbf{C}_k$ centroids, the main difference resides in the assignment and creation of the hash code. Each centroid is associated to a specific bit of the hash code:

\begin{equation}\label{eq:association}
\begin{cases}
	\parallel x-C_j \parallel \leq \delta   & j^{th}\; bit=1 \\ 
	\parallel x-C_j \parallel > \delta &  j^{th}\; bit=0
\end{cases}
\end{equation}

\noindent where $x$ is the feature point and $\delta$ is a threshold measure given by

\begin{equation}\label{eq:threshold}
\delta=\begin{cases} 
	(\prod_{j=1}^k\parallel x-C_j \parallel)^\frac{1}{k} \;\;\; \; \; \;   geometric\; mean \\
	\frac{1}{k} \sum_{j=1}^k \parallel x-C_j \parallel      \;\;\; \; \; \;   arithmetic\; mean \\
	n^{th} \;nearest \;distance \;\parallel x-C_j \parallel \;\;\;\forall j=1,...,k
\end{cases}
\end{equation}

\noindent i.e.~centroid $j$ is associated to the $j^{th}$ bit of the hash code of length $k$; the bit is set to $1$ if the feature to be quantized is assigned to its centroid, or to $0$ otherwise.


\medskip
A feature can be assigned to more than one centroid using two different approaches:

 \emph{i)} \textit{m-k-means-t} - using Eq.~(\ref{eq:association}) and one of the first two thresholds of Eq.~(\ref{eq:threshold}). In this case the feature vector is considered as belonging to all the centroids from which its distance is below the threshold. Experiments have shown that the arithmetic mean is more efficient with respect to the geometric one, and all the experiments will report results obtained with it.

\emph{ii)} \textit{m-k-means-n} - using Eq.~(\ref{eq:association}) and the third threshold of Eq.~(\ref{eq:threshold}), i.e.~assigning the feature to a predefined number $n$ of nearest centroids. 

We also introduce a variant (\textit{m-k-means-t2} and \textit{m-k-means-n2}) to the previous approaches by randomly splitting the training data into two groups and creating two different codebooks for each feature vector. The final hash code is given by the union of these two codes.

\begin{figure}[!h]
\begin{center}
\begin{minipage}[c]{.95\columnwidth}
\centering
\includegraphics[width=1\columnwidth]{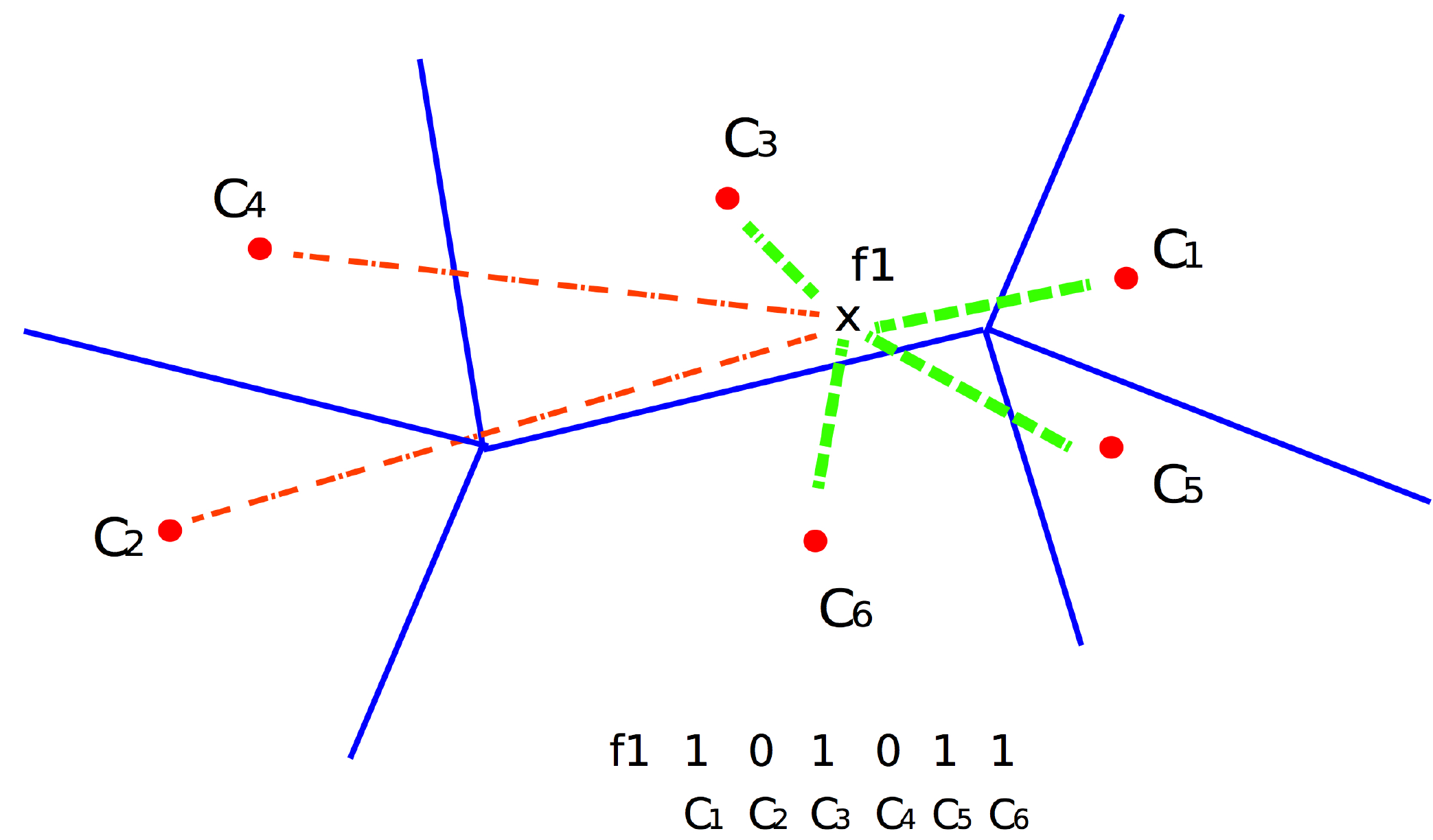}
\end{minipage}%
\hspace{10mm}%
\begin{minipage}[c]{.95\columnwidth}
\centering
\includegraphics[width=1\columnwidth]{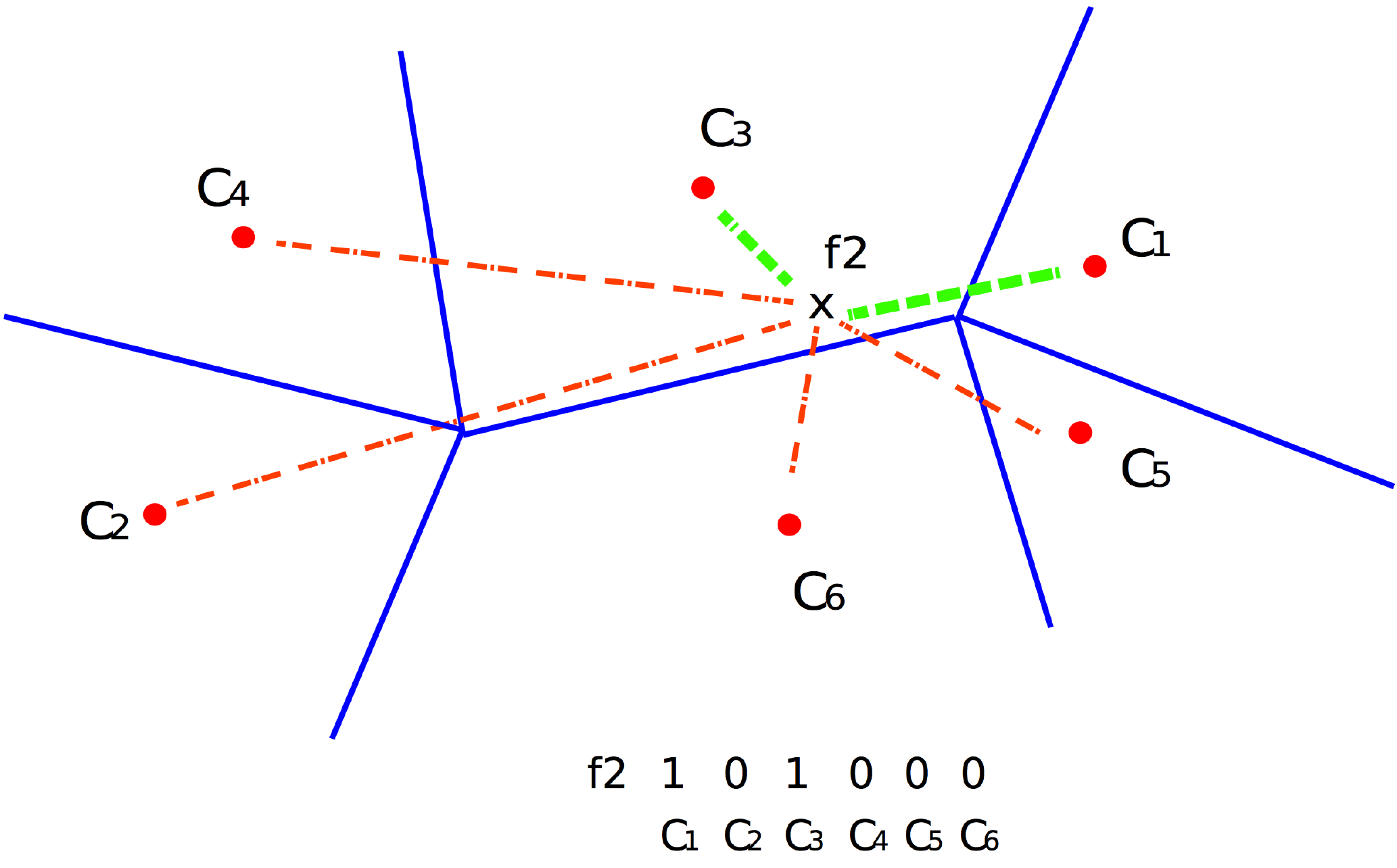}
\end{minipage}
\vspace{4mm}

\begin{minipage}[b]{.95\columnwidth}
\includegraphics[width=1\columnwidth]{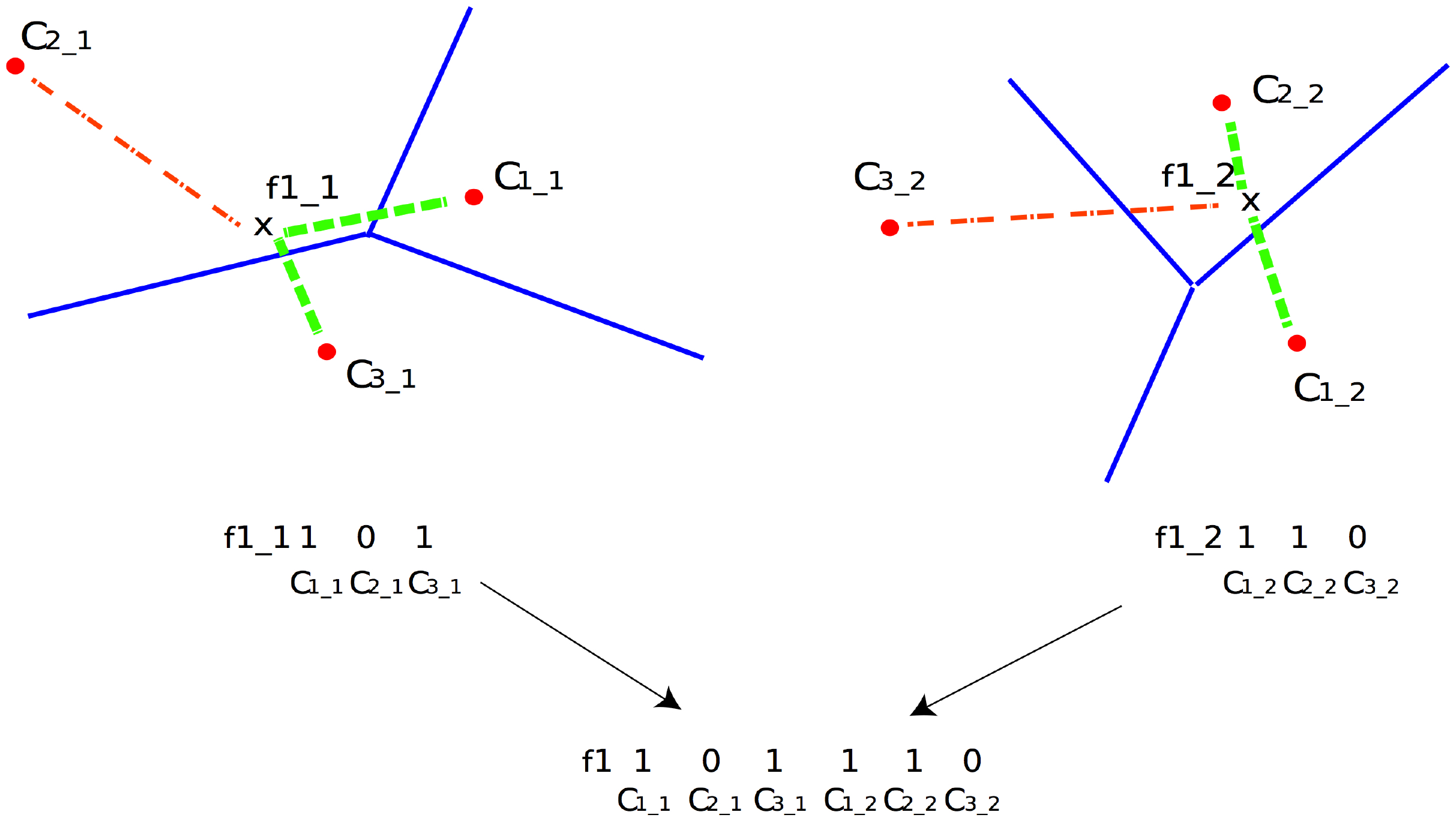}
\end{minipage}
\caption{Toy examples illustrating the proposed method: \emph{(top)} features can be assigned (green line) to a variable number of nearest clusters (e.g.~those with distances below the mean $\delta$ - i.e.~\emph{m-k-means-t}); \emph{(middle)} features can be assigned to a fixed number of clusters (e.g.~the 2 nearest clusters - i.e.~\emph{m-k-means-n});\emph{(bottom)} hash code created from two different codebooks (\emph{m-k-means-x2}, where $x$ can be either $t$ or $n$).
If a feature is assigned to a centroid the corresponding bit in the hash code is set to 1.}
\label{fig:quantization-example}
\end{center}
\end{figure}

With the proposed approach it is possible to create hash signatures using a much smaller number of centroids than using the usual k-means baseline, since each centroid is directly associated to a bit of the hash code. This approach can be considered a quantized version of codebook soft assignment \cite{vangemert-2010} and, similarly, it alleviates the problem of codeword ambiguity while reducing the quantization error.


%

\smallskip
Fig.~\ref{fig:quantization-example} illustrates the quantization process and the resulting hash codes in three cases: one in which a vector is assigned to a variable number of centroids (\emph{m-k-means-t}), one in which a vector is assigned to a predefined number of centroids (\emph{m-k-means-n}) and one in which the resulting code is created by the union of two different codes created using two different codebooks (\emph{m-k-means-x2}, where $x$ can be either $t$ or $n$). In all cases the feature is assigned to more than one centroid. An evaluation of these two approaches is reported in Sect.~\ref{sec:experiments}.

\smallskip
Typically a multi probe approach is used to solve the problem of ambiguous assignment to a codebook centroid (in case of vector quantization) or quantization error (e.g.~in case of scalar quantization). With this approach one or more bits of the query hash code are flipped to perform a query expansion, improving recall at the expense of computational cost and search time. In the proposed method this need of multi probe queries is greatly reduced, because of the possibility of assignment of features to more than one centroid. In fact, all the experiments have been performed without using multi probe querying.

\section{Experimental results}\label{sec:experiments}
The variants of the proposed method (\emph{m-k-means-t}, \emph{m-k-means-n}, \emph{m-k-means-t2} and \emph{m-k-means-n2}) have been thoroughly compared to several state-of-the-art approaches using standard datasets, experimental setups and evaluation metrics.

\subsection{Datasets}\label{sec:dataset}

\begin{figure}[!h]
\centering
\includegraphics[width=1\columnwidth]{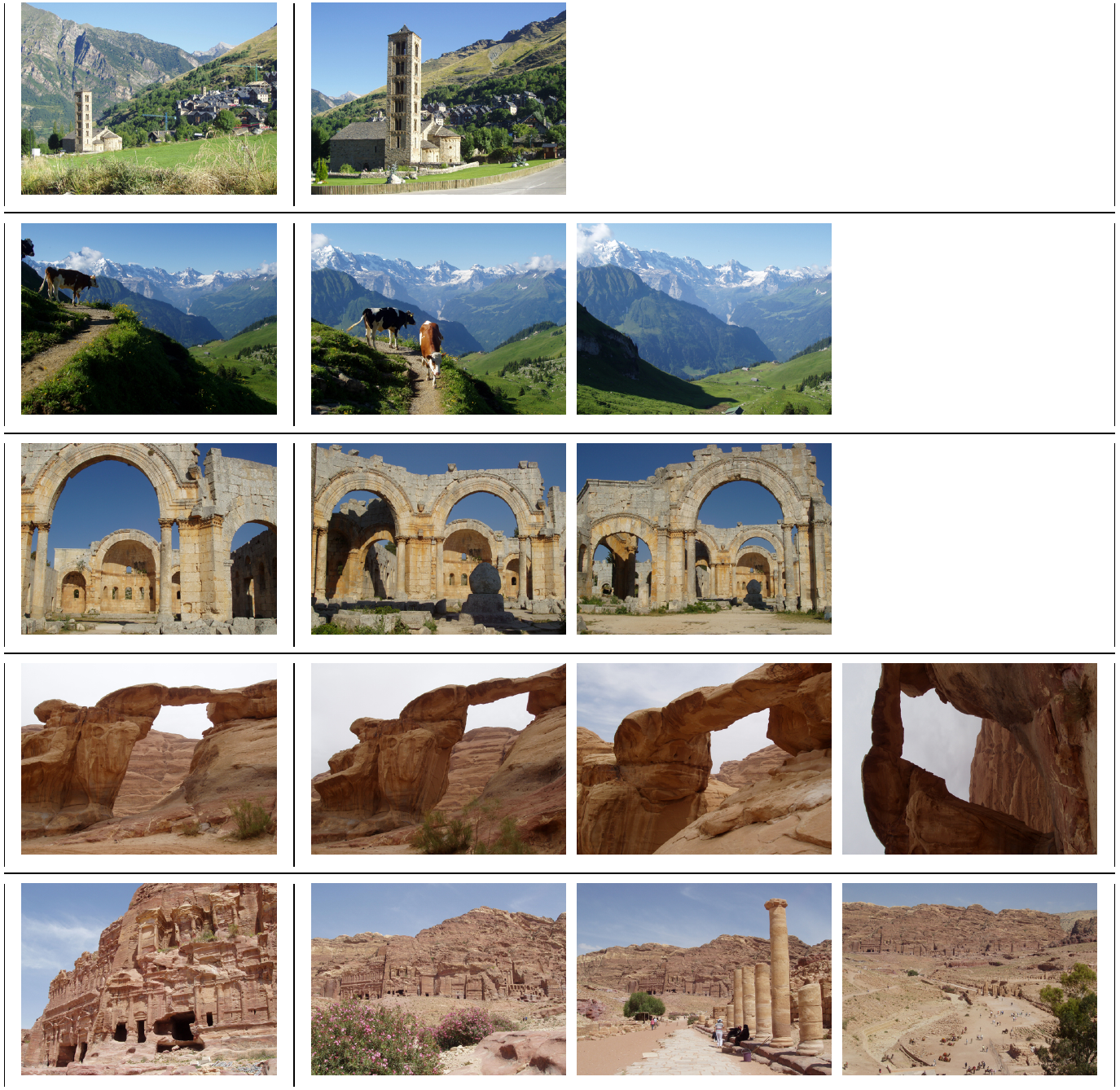}
\caption{Sample images from INRIA Holiday dataset. Left column shows the query images, the other columns show similar images.}
\label{fig:inria}
\end{figure}

\textbf{BIGANN Dataset} \cite{jegou-2011b, jegou-2011} is a large-scale  dataset commonly used to compare methods for visual feature hashing and approximate nearest neighbor search \cite{jegou-2011b, jegou-2011, babenko-2012, norouzi-2014, norouzi-2013, kalantidis2014locally, ge-2013}. The dataset is composed by three different sets of SIFT and GIST descriptors, each one divided in three subsets: a learning set, a query set and base set; each query has corresponding ground truth results in the base set, computed in an exhaustive way with Euclidean distance, ordered from the most similar to the most different. For SIFT1M and SIFT1B query and base descriptors have been extracted from the INRIA Holidays images \cite{jegou2008hamming}, while the learning set has been extracted from Flickr images. For GIST1M query and base descriptors are from INRIA Holidays and Flickr 1M datasets, while learning vectors are from \cite{torralba-2008-2}. In all the cases query descriptors are from the query images of INRIA Holidays (see Figure \ref{fig:inria}).
The characteristics of the dataset are summarized in Table \ref{tab:dataset}.

\begin{table}[!bh]
\centering
\caption{BIGANN datasets characteristics}\label{tab:dataset}
\resizebox{1\columnwidth}{!}{
\begin{tabular}{ l r r r}
 \hline

 \textbf{vector dataset}			& 	SIFT 1M 	& SIFT 1B 	&	GIST 1M\\
 \hline
 
 descriptor dimensionality $D$ 	& 	128		&    128   	&	960 \\

 \# learning set vectors 			&   100,000	&  100,000,000  &	500,000\\
 
 \# database set vectors 			&   1,000,000 &  1,000,000,000  &	1,000,000\\
 
 \# queries set vectors 			&  	10,000	&  10,000  	&	1,000\\
 
 \# nearest vectors for each query	&	100	&	1000		&	100\\
 
 \hline
 
\end{tabular}
}
\end{table}

\begin{figure}[!htb]
\centering
\includegraphics[width=1\columnwidth]{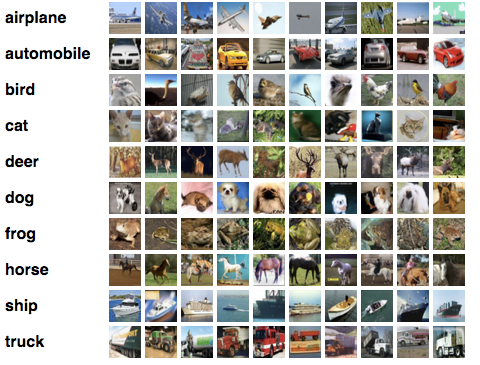}
\caption{Sample images from CIFAR-10 dataset}
\label{fig:cifar10}
\end{figure}

\textbf{CIFAR-10 Dataset} \cite{krizhevsky2009learning} consists of 60,000 colour images ($32 \times 32$ pixels) in 10 classes, with 6,000 images per class (see Figure \ref{fig:cifar10}). The dataset is split into training and test sets, composed by 50,000 and 10,000 images respectively. A retrieved image is considered relevant for the query if it belongs to the same class. This dataset has been used for ANN retrieval in \cite{xia2014supervised, lin2015deep}.
\medskip

\begin{figure}[!hbt]
\centering
\includegraphics[width=1\columnwidth]{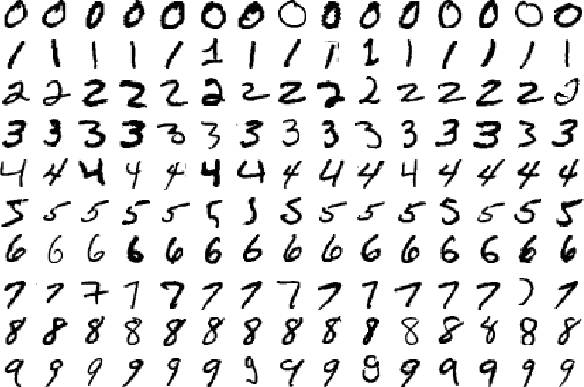}
\caption{Sample images from MNIST dataset}
\label{fig:mnist}
\end{figure}

\textbf{MNIST Dataset} \cite{lecun1998mnist} consists of 70,000 handwritten digits images ($28 \times 28$ pixels, see Figure \ref{fig:mnist}). The dataset is split into 60,000 training examples and 10,000 test examples. Similarly to CIFAR-10 a retrieved image is considered relevant if it belongs to the same class of the query. This dataset has been used for ANN retrieval in \cite{xia2014supervised, lin2015deep}.

\subsection{Evaluation Metrics}\label{sec:evaluation}

The performance of ANN retrieval in BIGANN dataset is evaluated using \textit{recall@R}, which is used in most of the results reported in the literature \cite{jegou-2011, jegou-2011b, norouzi-2013, kalantidis2014locally, ge-2013, babenko-2012}  and it is, for varying values of R,  the average rate of queries for which the 1-nearest neighbor is retrieved in the top R positions. In case of $R=1$ this metric coincides with \textit{precision@1}.

Performance of image retrieval in CIFAR-10 and MNIST is measured following the setup of \cite{xia2014supervised}, using Mean Average Precision:

\begin{equation}\label{eq:map}
MAP=\frac{\sum_{q=1}^QAveP(q)}{Q}
\end{equation}

\noindent where 

\begin{equation}\label{eq:avep}
AveP=\int_{0}^{1}p(r)dr 
\end{equation}

\noindent is the area under the precision-recall curve and $Q$ is the number of queries.

\subsection{Configurations and Implementations}\label{sec:config}


\paragraph{BIGANN} We use settings which reproduce top performances at 64-bit codes.
We perform search with a non-exhaustive approach. For each query 64 bits binary hash code of the feature and Hamming distance measure are used to extract small subsets of candidate from the whole database set (Table \ref{tab:dataset}). Euclidean distance measure is then used to re-rank the nearest feature points, calculating recall@R values in these subsets.

\begin{figure*}[!htb]
\centering
\includegraphics[width=1.5\columnwidth]{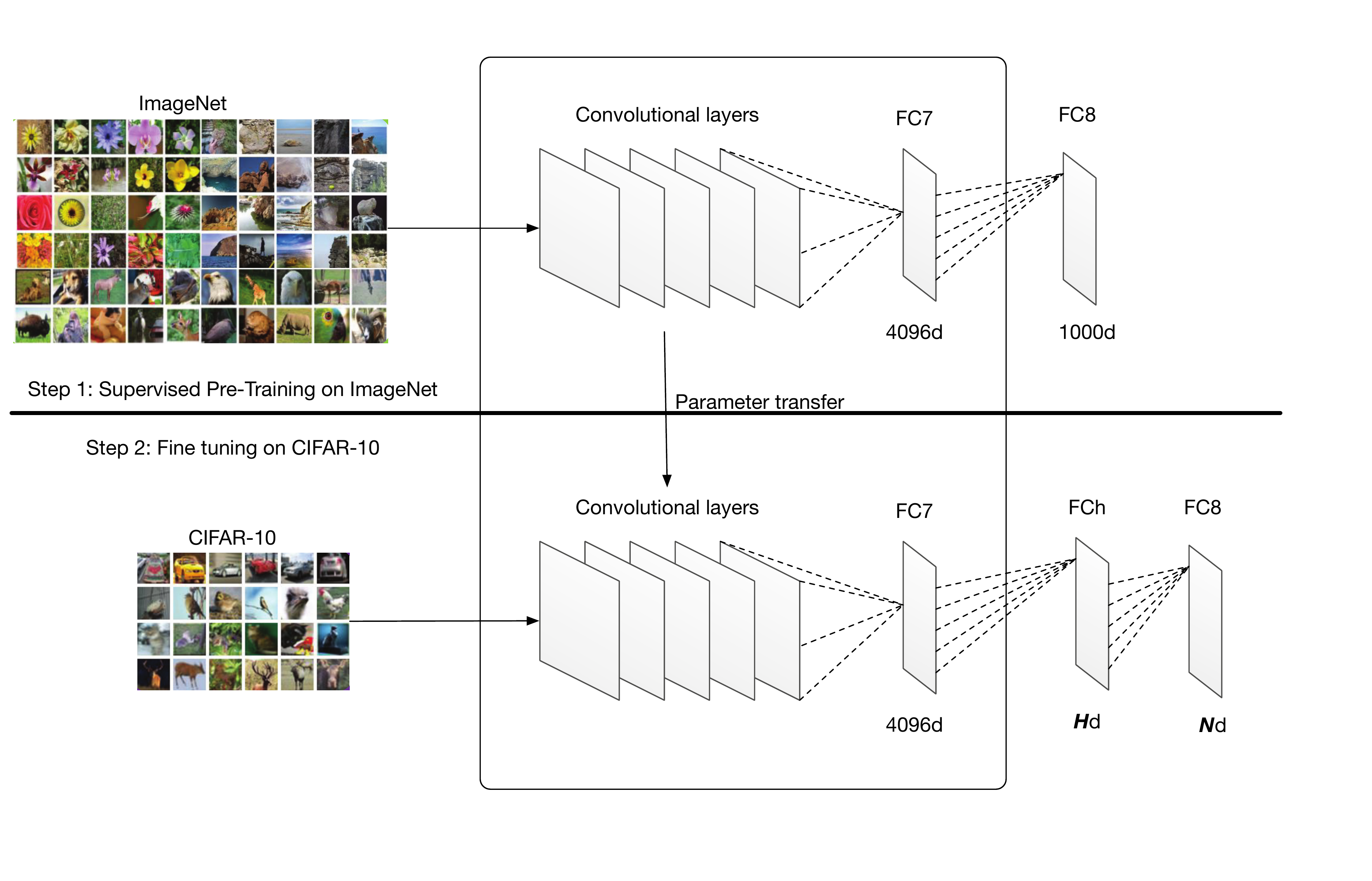}
\caption{Framework used for CNN feature extraction on CIFAR-10 \cite{lin2015deep}: we use the values of the nodes of the FCh layer as feature (48 dimensions).}
\label{fig:net}
\end{figure*}

\paragraph{CIFAR-10} We use features computed with the framework proposed in \cite{lin2015deep} (Figure \ref{fig:net}). The process is carried out in two steps: in the first step a supervised pre-training on the large-scale ImageNet dataset \cite{krizhevsky2012imagenet} is performed. In the second step fine-tuning of the network is performed, with the insertion of a latent layer that simultaneously learns domain specific feature representations and a set of hash-like functions. The authors used the pre-trained CNN model proposed by Krizhevsky \etal\cite{krizhevsky2012imagenet} and implemented in the Caffe CNN library \cite{jia2014caffe}.
In our experiments we use features coming from the $FCh$ Layer (Latent Layer H), which has a size of 48 nodes.

\paragraph{MNIST} We use LeNet CNN to compute our features in MNIST. This is a network architecture developed by LeCun \cite{lecun1998gradient} that was especially designed for recognizing handwritten digits, reading zip codes, etc.
It is a 8-layer network with 1 input layer, 2 convolutional layers, 2 non-linear down-sampling layers, 2 fully connected layers and a Gaussian connected layer with 10 output classes. We used a modified version of LeNet \cite{jia2014caffe} and we obtain features from the first fully connected layer.

\medskip
We perform search with a non-exhaustive approach on both CIFAR-10 and MNIST datasets. For each image we extract a 48-dimensional feature vector for CIFAR-10, and 500-dimensional feature vector for MNIST, from the respective network and then we generate a 48 bits binary hash code using the proposed methods of Sect.~\ref{sec:method}. Hamming distance is used to select the nearest hash codes for each query and similarity measure given by

\begin{equation}\label{eq:avep}
similarity=\cos(\theta)=\frac{\textbf{A}\cdot\textbf{B}}{\parallel\textbf{A}\parallel\parallel\textbf{B}\parallel}=\frac{\sum_{i=1}^nA_iB_i}{\sqrt{\sum_{i=1}^n}A^2_i \sqrt{\sum_{i=1}^n}B^2_i}
\end{equation}

where $A_i$ and $B_i$ are the components of the original feature vectors $A$ and $B$, is used to re-rank the nearest visual features.

\subsection{Results on BIGANN: SIFT1M, GIST1M}\label{sec:comp1}

In this set of experiments the proposed approach and its variants are compared on the SIFT1M (Table \ref{tab:comparison_SIFT1M}) and GIST1M (Table \ref{tab:comparison_GIST1M}) datasets against four methods discussed in section \ref{sec:previous-work}: Product Quantization  (ADC and IVFADC)\cite{jegou-2011}, Cartesian k-means \cite{norouzi-2013}, LOPQ \cite{kalantidis2014locally} and a non-exhaustive adaptation of OPQ \cite{ge-2013}, called I-OPQ \cite{kalantidis2014locally}.


ADC (\textit{Asymmetric Distance Computation}) is characterized by the number of sub vectors \textit{m} and the number of quantizers per sub vectors $k^{*}$, and produces a code of length \textit{m} $\times log_{2}k^{*}$.
IVFADC (\textit{Inverted File with Asymmetric Distance Computation}) is characterized by the codebook size $k^{\prime}$ (number of centroids associated to each quantizer), the number of neighbouring cells \textit{w} visited during the multiple assignment, the number of sub vectors \textit{m} and  the number of quantizers per sub vectors $k^{*}$ which is in this case fixed to $k^{*}=256$. The length of the final code is given by \textit{m} $\times log_{2}k^{*}$.

Cartesian k-means (ck-means) \cite{norouzi-2013} models region center as an additive combinations of subcenters. Let $m$ be the number of subcenters, with $h$ elements, then the total number of model centers is $k=h^m$, but the total number of subcenters is $h \times m$, and the number of bits of the signature is $m \times log_{2} h$.

 LOPQ (\textit{The Locally optimized product quantization}) \cite{kalantidis2014locally} is a vector quantizer that combines low distortion with fast search applying a local optimization over rotation and space decomposition.

I-OPQ \cite{kalantidis2014locally} is  a non-exhaustive adaptation of OPQ (\textit{Optimized Product Quantization}) \cite{ge-2013} which use either OPQ${_{P}}$  or  OPQ${_{NP}}$ global optimization.

The parameters of the proposed methods are set as follows: for \textit{m-k-means-t} we use as  threshold the arithmetic mean of the distances between feature vectors and centroids to compute hash code; \textit{m-k-means-n} creates hash code by setting to 1 the corresponding position of the first 32 (SIFT1M) and first 48 (GIST1M) nearest centroids for each feature; \textit{m-k-means-t2} and \textit{m-k-means-n2} create two different sub hash codes for each feature by splitting into two parts the training phase and combine these two sub parts into one single code to create the final signature. Since we have a random  splitting during the training phase, these experiments are averaged over a set of 10 runs.

\begin{table}[!bth]
\centering
\caption{\textit{Recall@R} on SIFT1M - Comparison between our method (m-k-means-t, m-k-means-n with n=32, m-k-means-t2 and m-k-means-n2 with n=32), the Product Quantization method (ADC and IVFADC) \cite{jegou-2011}, Cartesian k-means method (cm-means) \cite{norouzi-2013}, a non-exhaustive adaptation of the Optimized Product Quantization method (I-OPQ) and a Locally optimized product quantization method (LOPQ) \cite{kalantidis2014locally}.}  \label{tab:comparison_SIFT1M} 

\resizebox{1\columnwidth}{!}{
\begin{tabular}{ |p{2.8cm}|c| c | c | c| c| c| c| }

 \hline
method		 						& 	R@1   	& 	R@10    &	R@100	&	R@1000	&	R@10000	\\
 \hline

ADC \cite{jegou-2011}				&	0.224	&  	0.600	& 	0.927	& 	0.996 	& 	0.999  	\\
  
IVFADC \cite{jegou-2011}				& 	0.320	&  	0.739 	& 	0.953	& 	0.972 	& 	0.972  	\\
 
ck-means \cite{norouzi-2013}			&	0.231	& 	0.635 	& 	0.930	& 	\textbf{1} 	&	\textbf{1}	\\
 
I-OPQ \cite{kalantidis2014locally} 	&   0.299   & 	0.691   & 	0.875  	& 	0.888  	& 	0.888  	\\
 
LOPQ \cite{kalantidis2014locally}  	&   0.380   & 	0.780  	& 	0.886 	& 	0.888   & 	0.888  	\\
  
\textbf{m-k-means-t }   	&   0.501 & 0.988 & \textbf{1} & \textbf{1} & \textbf{1} \\
\textbf{m-k-means-t2 }   				&   \textbf{0.590} & \textbf{0.989} & \textbf{1} & \textbf{1} & \textbf{1} \\

\textbf{m-k-means-n }   				& 0.436   & 0.986 & \textbf{1} & \textbf{1} & \textbf{1}  \\
\textbf{m-k-means-n2 }   				& 0.561   & 0.986 & \textbf{1} & \textbf{1} & \textbf{1}  \\
   
 \hline
\end{tabular}
}
\end{table}

\begin{table}[!htb]
\centering
\caption{\textit{Recall@R} on GIST1M - Comparison between our method (m-k-means-t, m-k-means-n with n=48, m-k-means-t2 and m-k-means-n2 with n=48), the Product Quantization method (ADC and IVFADC) \cite{jegou-2011}, Cartesian k-means method (ck-means) \cite{norouzi-2013}, a non-exhaustive adaptation of the Optimized Product Quantization method (I-OPQ)\cite{kalantidis2014locally} and a Locally optimized product quantization method (LOPQ) \cite{kalantidis2014locally}.}\label{tab:comparison_GIST1M}
\resizebox{1\columnwidth}{!}{
\begin{tabular}{ |p{2.5cm}|c | c | c | c | c | c | c | }

 \hline
method		 						& 	R@1   	& 	R@10    & 	R@100   &	R@1000	&  R@10000		\\
 \hline

ADC\cite{jegou-2011}				&	0.145	&  	0.315  	& 	0.650	& 	0.932 	& 0.997  		\\
 
IVFADC\cite{jegou-2011}				& 	0.180	&  	0.435 	&   0.740  	& 	0.966 	& 0.992  		\\
 
ck-means\cite{norouzi-2013}			&	0.135	& 	0.335 	& 	0.728	& 	0.952   & 0.985   		\\
 
I-OPQ\cite{kalantidis2014locally}   &   0.146   & 	0.410  	& 	0.729 	& 	0.862   & 0.866  		\\
 
LOPQ\cite{kalantidis2014locally}    &   0.160  	& 	0.461   & 	0.756 	& 	0.860   & 0.866  		\\
  
\textbf{m-k-means-t }   				&   0.111 	& 0.906 & \textbf{1} & \textbf{1}  & \textbf{1} 	\\

\textbf{m-k-means-t2 }   				&   0.123 	& 0.890 & \textbf{1} & \textbf{1}  & \textbf{1} 	\\

\textbf{m-k-means-n}   				&   0,231	& \textbf{0,940 }& \textbf{1}  &  \textbf{1}  & \textbf{1}	\\

\textbf{m-k-means-n2}   				&   \textbf{0,265}	& 0,905 & \textbf{1}  &  0,999  & 0,999	\\
   
 \hline
 
\end{tabular}

}
\end{table}

The proposed method obtains the best results when considering the more challenging values of \textit{recall@R}, i.e.~with a small number of nearest neighbors, like 1, 10 and 100. When R goes to 1000 and 10,000 it still obtains the best results and in the case of SIFT1M it is on par with ck-means \cite{norouzi-2013}. Considering GIST1M the method consistently outperforms all the other methods for all the values of R.

\subsection{Results on BIGANN: SIFT1B}\label{sec:comp2}
In this experiment we compare our method on the SIFT1B dataset (Table \ref{tab:comparison_SIFT1B}) against LOPQ and a sub-optimal variant LOR+PQ \cite{kalantidis2014locally}, a single index approaches IVFADC \cite{jegou-2011}, I-OPQ \cite{ge-2013} and a multi-index method Multi-D-ADC \cite{babenko-2012}. 
m-k-means-t uses the same setup of the previous experiment.

Also in this case the proposed method obtains the best results, in particular when considering the more challenging small values of $R$ for the \textit{recall@R} measure ($R=1$ and $R=10$), with an improvement between $3\times$ and $1.5 \times$ the best results of compared methods.

\begin{table}[!htb]
\centering
\caption{\textit{Recall@R} on SIFT1B - Comparison between our method (m-k-means-t), the Product Quantization method \cite{jegou-2011b}, a non-exhaustive adaptation of the Optimized Product Quantization method  (I-OPQ), a multi-index method (Multi-D-ADC), a Locally optimized product quantization method (LOPQ) with a sub-optimal variant (LOR+PQ)}\label{tab:comparison_SIFT1B}
\resizebox{1\columnwidth}{!}{
\begin{tabular}{ |p{2.8cm}|c | c | c | c | c | c | c | }

 \hline
method		 						& 	R@1   	& 	R@10     & 	R@100   \\
 \hline

IVFADC\cite{jegou-2011b} 			& 	0.088	&  	0.372 	& 	0.733	\\
 
IVFADC\cite{jegou-2011}	 			& 	0.106 	&  	0.379 	& 	0.748 	\\
 
ck-means\cite{norouzi-2013}	 		&	0.084	& 	0.288 	& 	0.637  	\\
 
I-OPQ\cite{kalantidis2014locally}	&   0.114   & 	0.399   & 	0.777  	\\
 
Multi-D-ADC\cite{babenko-2012}      &   0.165   & 	0.517   & 	0.860  	\\

LOR+PQ\cite{kalantidis2014locally}  &   0.183   & 	0.565   & 	0.889  	\\
 
LOPQ\cite{kalantidis2014locally}    &   0.199  & 	0.586   & 	0.909  	\\
  
\textbf{m-k-means-t }   				&   \textbf{0.668} 	& \textbf{0.893} & \textbf{0.926}  \\
   
 \hline
\end{tabular}
}
\end{table}

\subsection{Results on CIFAR-10, MNIST}\label{sec:comp3}

In the experiments on CIFAR-10 \cite{krizhevsky2009learning} and MNIST \cite{lecun1998mnist}  images dataset we use the the following configurations for the proposed method: hash code length of 48 bits (the same length used by the compared methods), arithmetic mean for the \emph{m-k-means-t} variant, $n=24$ for \emph{m-k-means-n}. 

Queries are performed using a random selection of 1,000 query images (100 images for each class), considering a category labels ground truth relevance ($Rel(i)$) between a query $q$ and the $i^{th}$ ranked image. So $Rel(i) \in \left\{0,1\right\} $ with 1 for the query and $i^{th}$ images with the same label and 0 otherwise.  This setup has been used in \cite{lin2015deep, xia2014supervised}.
Since we select queries in a random way the results of these experiments are averaged over a set of 10 runs.

\begin{table}[!tbh]
\centering
\caption{MAP results on CIFAR-10 and MNIST. Comparison between our method (m-k-means-t, m-k-means-n with n=24, m-k-means-t2 and m-k-means-n2 with n=24) with KSH \cite{liu2012supervised}, ITQ-CCA  \cite{gong-2011}, MLH \cite{norouzi2011minimal}, BRE \cite{kulis2009learning}, CNNH \cite{xia2014supervised}, CNNH+ \cite{xia2014supervised}, KevinNet \cite{lin2015deep}, LSH \cite{gionis-1999}, SH \cite{weiss-2009}, ITQ \cite{gong-2011}}\label{tab:comparison_CIFAR10_MNIST} \label{tab:comparison_CIFAR10_MNIST}
\resizebox{1 \columnwidth}{!}{
\begin{tabular}{ |p{2.5cm}|c | c |  }

 \hline
method		 			& 	CIFAR-10 (MAP)   	&   MNIST (MAP)   \\
 \hline

LSH \cite{gionis-1999} 			& 	0.120	&	0.243	\\

SH \cite{weiss-2009}	 		&	0.130	&	 0.250	\\
 
ITQ\cite{gong-2011}	 			& 	0.175 	& 	0.429 	\\
 
BRE \cite{kulis2009learning}	&   0.196    & 0.634 	\\
 
MLH \cite{norouzi2011minimal}      &   0.211  &  0.654   	\\

ITQ-CCA\cite{gong-2011}  &   0.295    &  0.726 	\\
 
KSH \cite{liu2012supervised}   &   0.356  &   0.900  	\\

CNNH \cite{xia2014supervised}   &   0.522  &   0.960 	\\

CNNH+ \cite{xia2014supervised}   &   0.532  &   0.975 	\\

KevinNet \cite{lin2015deep}   &   0.894 &  \textbf{0.985} \\
  
\textbf{m-k-means-t }   		& 0.953	&   0.972	 \\

\textbf{m-k-means-t2 }   		& 0.849  	&  0.964 	 \\

\textbf{m-k-means-n }   		&  \textbf{0.972}  	&  0.969 	 \\

\textbf{m-k-means-n2 }   		& 0.901  	&   	0.959 \\

 \hline
\end{tabular}
}
\end{table}

We compared the proposed approach with several state-of-the-art hashing methods including supervised (KSH \cite{liu2012supervised}, ITQ-CCA  \cite{gong-2011}, MLH \cite{norouzi2011minimal}, BRE \cite{kulis2009learning}, CNNH \cite{xia2014supervised}, CNNH+ \cite{xia2014supervised}, KevinNet \cite{lin2015deep}) and unsupervised methods (LSH \cite{gionis-1999}, SH \cite{weiss-2009}, ITQ \cite{gong-2011}).

The proposed method obtains the best results on the more challenging of the two datasets, i.e.~CIFAR-10. The comparison with the KevinNet \cite{lin2015deep} method is interesting since we use the same features, but we obtain better results for all the variants of the proposed method except one.
On MNIST dataset the best results of our approach are comparable with the second best method \cite{xia2014supervised}, and anyway are not far from the best approach \cite{lin2015deep}.

\section{Conclusions}\label{sec:conclusions}
We have proposed a new version of the k-means based hashing schema called multi-k-means -- with 4 variants: \emph{m-k-means-t}, \emph{m-k-means-t2}, \emph{m-k-means-n} and \emph{m-k-means-n2} -- which uses a small number of centroids, guarantees a low computational cost and results in a compact quantizer. Our compact hash signature is able to represent high dimensional visual features obtaining a very high efficiency in approximate nearest neighbor (ANN) retrieval. The method has been tested on large scale datasets of engineered (SIFT and GIST) and learned (deep CNN) features, obtaining results that outperform or are comparable to more complex state-of-the-art approaches.

\hide{
\section{Acknowledgments} 
This work is partially supported by the ``Social Museum and Smart Tourism'' project (CTN01\_00034\_231545).

This research is based upon work supported [in part] by the Office of the Director of National Intelligence (ODNI), Intelligence Advanced Research Projects Activity (IARPA), via IARPA contract number 2014-14071600011. The views and conclusions contained herein are those of the authors and should not be interpreted as necessarily representing the official policies or endorsements, either expressed or implied, of ODNI, IARPA, or the U.S. Government.  The U.S. Government is authorized to reproduce and distribute reprints for Governmental purpose notwithstanding any copyright annotation thereon.
}

%
\bibliographystyle{abbrv}


\begin{thebibliography}{10}

\bibitem{arthur2007k}
D.~Arthur and S.~Vassilvitskii.
\newblock k-means++: The advantages of careful seeding.
\newblock In {\em Proc. of ACM-SIAM symposium on Discrete algorithms (SODA)},
  2007.

\bibitem{babenko-2012}
A.~Babenko and V.~Lempitsky.
\newblock The inverted multi-index.
\newblock In {\em Proc. of IEEE Conference on Computer Vision and Pattern
  Recognition (CVPR)}, 2012.

\bibitem{boyd2011distributed}
S.~Boyd, N.~Parikh, E.~Chu, B.~Peleato, and J.~Eckstein.
\newblock Distributed optimization and statistical learning via the alternating
  direction method of multipliers.
\newblock {\em Foundations and Trends in Machine Learning}, 3(1):1--122, 2011.

\bibitem{chandrasekhar-2010}
V.~Chandrasekhar, M.~Makar, G.~Takacs, D.~Chen, S.~S. Tsai, N.-M. Cheung,
  R.~Grzeszczuk, Y.~Reznik, and B.~Girod.
\newblock Survey of {SIFT} compression schemes.
\newblock In {\em Proc. of International Workshop on Mobile Multimedia
  Processing (WMPP)}, 2010.

\bibitem{Chatfield14}
K.~Chatfield, K.~Simonyan, A.~Vedaldi, and A.~Zisserman.
\newblock Return of the devil in the details: Delving deep into convolutional
  nets.
\newblock In {\em Proc. of British Machine Vision Conference (BMVC)}, 2014.

\bibitem{chen-2015}
C.-C. Chen and S.-L. Hsieh.
\newblock Using binarization and hashing for efficient {SIFT} matching.
\newblock {\em Journal of Visual Communication and Image Representation},
  30:86--93, 2015.

\bibitem{do2015discrete}
T.-T. Do, A.-Z. Doan, and N.-M. Cheung.
\newblock Discrete hashing with deep neural network.
\newblock {\em arXiv preprint arXiv:1508.07148}, 2015.

\bibitem{du-2014}
S.~Du, W.~Zhang, S.~Chen, and Y.~Wen.
\newblock Learning flexible binary code for linear projection based hashing
  with random forest.
\newblock In {\em Proc. of International Conference on Pattern Recognition
  (ICPR)}, 2014.

\bibitem{ge-2013}
T.~Ge, K.~He, Q.~Ke, and J.~Sun.
\newblock Optimized product quantization for approximate nearest neighbor
  search.
\newblock In {\em Proc. of IEEE Conference on Computer Vision and Pattern
  Recognition (CVPR)}, 2013.

\bibitem{gionis-1999}
A.~Gionis, P.~Indyk, and R.~Motwani.
\newblock Similarity search in high dimensions via hashing.
\newblock In {\em Proc. of International Conference on Very Large Data Bases
  (VLDB)}, 1999.

\bibitem{gong-2011}
Y.~Gong and S.~Lazebnik.
\newblock Iterative quantization: A procrustean approach to learning binary
  codes.
\newblock In {\em Proc. of IEEE Conference on Computer Vision and Pattern
  Recognition (CVPR)}, 2011.

\bibitem{guo2015cnn}
J.~Guo and J.~Li.
\newblock {CNN} based hashing for image retrieval.
\newblock {\em arXiv preprint arXiv:1509.01354}, 2015.

\bibitem{he2013k}
K.~He, F.~Wen, and J.~Sun.
\newblock K-means hashing: An affinity-preserving quantization method for
  learning binary compact codes.
\newblock In {\em Proc. of IEEE Conference on Computer Vision and Pattern
  Recognition (CVPR)}, 2013.

\bibitem{heo-2012}
J.-P. Heo, Y.~Lee, J.~He, S.-F. Chang, and S.-E. Yoon.
\newblock Spherical hashing.
\newblock In {\em Proc. of IEEE Conference on Computer Vision and Pattern
  Recognition (CVPR)}, 2012.

\bibitem{jain-2011}
M.~Jain, H.~J{\'e}gou, and P.~Gros.
\newblock Asymmetric {Hamming} embedding: Taking the best of our bits for large
  scale image search.
\newblock In {\em Proc. of ACM Multimedia (ACM MM)}, 2011.

\bibitem{jegou2008hamming}
H.~Jegou, M.~Douze, and C.~Schmid.
\newblock Hamming embedding and weak geometric consistency for large scale
  image search.
\newblock In {\em Proc. of European Conference on Computer Vision (ECCV)},
  2008.

\bibitem{jegou-2011}
H.~J\'egou, M.~Douze, and C.~Schmid.
\newblock Product quantization for nearest neighbor search.
\newblock {\em IEEE Transactions on Pattern Analysis and Machine Intelligence},
  33(1):117--128, 2011.

\bibitem{jegou-2011b}
H.~J\'egou, R.~Tavenard, M.~Douze, and L.~Amsaleg.
\newblock Searching in one billion vectors: re-rank with source coding.
\newblock In {\em Proc. of IEEE International Conference on Acoustics, Speech
  and Signal Processing (ICASSP)}, 2011.

\bibitem{jia2014caffe}
Y.~Jia, E.~Shelhamer, J.~Donahue, S.~Karayev, J.~Long, R.~Girshick,
  S.~Guadarrama, and T.~Darrell.
\newblock Caffe: Convolutional architecture for fast feature embedding.
\newblock In {\em Proc. of ACM Multimedia (ACM MM)}, pages 675--678, 2014.

\bibitem{jin-2014}
Z.~Jin, C.~Li, Y.~Lin, and D.~Cai.
\newblock Density sensitive hashing.
\newblock {\em IEEE Transactions on Cybernetics}, 44(8):1362--1371, 2014.

\bibitem{kalantidis2014locally}
Y.~Kalantidis and Y.~Avrithis.
\newblock Locally optimized product quantization for approximate nearest
  neighbor search.
\newblock In {\em Proc. of IEEE Conference on Computer Vision and Pattern
  Recognition (CVPR)}, pages 2329--2336, 2014.

\bibitem{krizhevsky2009learning}
A.~Krizhevsky and G.~Hinton.
\newblock Learning multiple layers of features from tiny images.
\newblock Master's thesis, Department of Computer Science, University of
  Toronto, 2009.

\bibitem{krizhevsky2012imagenet}
A.~Krizhevsky, I.~Sutskever, and G.~E. Hinton.
\newblock Imagenet classification with deep convolutional neural networks.
\newblock In {\em Proc. of Neural Information Processing Systems (NIPS)}, pages
  1097--1105, 2012.

\bibitem{kulis2009learning}
B.~Kulis and T.~Darrell.
\newblock Learning to hash with binary reconstructive embeddings.
\newblock In {\em Proc. of Neural Information Processing Systems (NIPS)}, pages
  1042--1050, 2009.

\bibitem{lecun1998gradient}
Y.~LeCun, L.~Bottou, Y.~Bengio, and P.~Haffner.
\newblock Gradient-based learning applied to document recognition.
\newblock {\em Proceedings of the IEEE}, 86(11):2278--2324, 1998.

\bibitem{lecun1998mnist}
Y.~LeCun, C.~Cortes, and C.~J. Burges.
\newblock The {MNIST} database of handwritten digits, 1998.

\bibitem{lin2015deep}
K.~Lin, H.-F. Yang, J.-H. Hsiao, and C.-S. Chen.
\newblock Deep learning of binary hash codes for fast image retrieval.
\newblock In {\em Proc. of IEEE Conference on Computer Vision and Pattern
  Recognition (CVPR)}, 2015.

\bibitem{liu2012supervised}
W.~Liu, J.~Wang, R.~Ji, Y.-G. Jiang, and S.-F. Chang.
\newblock Supervised hashing with kernels.
\newblock In {\em Proc. of IEEE Conference on Computer Vision and Pattern
  Recognition (CVPR)}, pages 2074--2081, 2012.

\bibitem{norouzi-2013}
M.~Norouzi and D.~Fleet.
\newblock Cartesian k-means.
\newblock In {\em Proc. of IEEE Conference on Computer Vision and Pattern
  Recognition (CVPR)}, 2013.

\bibitem{norouzi2011minimal}
M.~Norouzi and D.~J. Fleet.
\newblock Minimal loss hashing for compact binary codes.
\newblock In {\em Proc. of International Conference in Machine Learning
  (ICML)}, 2011.

\bibitem{norouzi-2014}
M.~Norouzi, A.~Punjani, and D.~Fleet.
\newblock Fast exact search in {Hamming} space with multi-index hashing.
\newblock {\em IEEE Transactions on Pattern Analysis and Machine Intelligence},
  36(6):1107--1119, 2014.

\bibitem{pauleve-2010}
L.~Paulev\'e, H.~J\'egou, and L.~Amsaleg.
\newblock Locality sensitive hashing: A comparison of hash function types and
  querying mechanisms.
\newblock {\em Pattern Recognition Letters}, 31(11):1348--1358, 2010.

\bibitem{ren-2014}
G.~Ren, J.~Cai, S.~Li, N.~Yu, and Q.~Tian.
\newblock Scalable image search with reliable binary code.
\newblock In {\em Proc. of ACM Multimedia (ACM MM)}, 2014.

\bibitem{torralba-2008-2}
A.~Torralba, R.~Fergus, and W.~T. Freeman.
\newblock 80 million tiny images: A large data set for nonparametric object and
  scene recognition.
\newblock {\em IEEE Transactions on Pattern Analysis and Machine Intelligence},
  30(11):1958--1970, 2008.

\bibitem{vangemert-2010}
J.~van Gemert, C.~Veenman, A.~Smeulders, and J.-M. Geusebroek.
\newblock Visual word ambiguity.
\newblock {\em IEEE Transactions on Pattern Analysis and Machine Intelligence},
  32(7):1271--1283, 2010.

\bibitem{weiss-2009}
Y.~Weiss, A.~Torralba, and R.~Fergus.
\newblock Spectral hashing.
\newblock In {\em Proc. of Neural Information Processing Systems (NIPS)}, 2009.

\bibitem{xia2014supervised}
R.~Xia, Y.~Pan, H.~Lai, C.~Liu, and S.~Yan.
\newblock Supervised hashing for image retrieval via image representation
  learning.
\newblock In {\em Proc. of AAAI Conference on Artificial Intelligence (AAAI)},
  2014.

\bibitem{zhang2015supervised}
Z.~Zhang, Y.~Chen, and V.~Saligrama.
\newblock Supervised hashing with deep neural networks.
\newblock {\em arXiv preprint arXiv:1511.04524}, 2015.

\bibitem{zhou-2012}
W.~Zhou, Y.~Lu, H.~Li, and Q.~Tian.
\newblock Scalar quantization for large scale image search.
\newblock In {\em Proc. of ACM Multimedia (ACM MM)}, 2012.

\end{thebibliography}
%
%

\balancecolumns 

\end{document}